\pdfoutput=1
\documentclass[11pt]{article}

\usepackage[preprint]{acl}

\usepackage{times}
\usepackage{latexsym}
\usepackage{hyperref}
\usepackage[T1]{fontenc}

\usepackage[utf8]{inputenc}

\usepackage{microtype}

\usepackage{inconsolata}

\usepackage{graphicx}

\usepackage{algorithm}
\usepackage[noend]{algpseudocode}
\usepackage{lineno}
\usepackage{tikz}
\usepackage[most]{tcolorbox}
\usepackage{hyperref}
\usepackage{url}
\usepackage{graphicx} % Required for inserting images
\usepackage{soul}
\usepackage{xcolor}
\usepackage{tcolorbox}
\usepackage{enumitem}
\usepackage{multirow}
\usepackage{booktabs}
\usepackage{bbm}
\usepackage{bm}
\usepackage{booktabs}
\usepackage{graphicx}
\usepackage{multirow}
\usepackage{multicol}
\usepackage{enumitem}

\usepackage{booktabs}
\usepackage{tabularx}
\usepackage{adjustbox}
\usepackage{xspace}
\usepackage{comment}

\newcommand{\method}{\texttt{Group Preference Alignment}\xspace}
\newcommand{\methodshort}{\texttt{GPA}\xspace}
\newcommand{\methodCT}{\texttt{GPA-CT}\xspace}
\newcommand{\methodFT}{\texttt{GPA-FT}\xspace}
\usepackage{amsmath,amsfonts,bm}

\def\eqref#1{equation~\ref{#1}}
\def\1{\bm{1}}

\DeclareMathAlphabet{\mathsfit}{\encodingdefault}{\sfdefault}{m}{sl}
\SetMathAlphabet{\mathsfit}{bold}{\encodingdefault}{\sfdefault}{bx}{n}

\def\gA{{\mathcal{A}}}

\def\gD{{\mathcal{D}}}
\def\gE{{\mathcal{E}}}

\def\gG{{\mathcal{G}}}

\def\gI{{\mathcal{I}}}
\def\gJ{{\mathcal{J}}}

\def\gR{{\mathcal{R}}}

\title{\textit{Group Preference Alignment}: Customized LLM \\Response Generation from In-Situ Conversations}

\author{
  \textbf{Ishani Mondal\textsuperscript{$^{*\ddagger}$}},
  \textbf{Jack W. Stokes\textsuperscript{$^{*\dagger}$}},
  \textbf{Sujay Kumar Jauhar\textsuperscript{$^{*\dagger}$}}, \\
  \textbf{Longqi Yang\textsuperscript{$^{\dagger}$}},
  \textbf{Mengting Wan\textsuperscript{$^{\dagger}$}},
  \textbf{Xiaofeng Xu\textsuperscript{$^{\dagger}$}},
  \textbf{Xia Song\textsuperscript{$^{\dagger}$}},
  \textbf{Jennifer Neville\textsuperscript{$^{\thanks{\small{Corresponding authors: imondal@umd.edu, [jstokes | sjauhar | jenneville]@microsoft.com.
  }
}\dagger}$}}
\vspace{2mm}
\\
  \textsuperscript{$^{\ddagger}$}University of Maryland, College Park, 
   \textsuperscript{$^{\dagger}$}Microsoft Research, Redmond
\\
}

\begin{document}
\maketitle
\begin{abstract}
LLMs often fail to meet the specialized needs of distinct user groups due to their \texttt{one-size-fits-all} training paradigm \cite{lucy-etal-2024-one} and there is limited research on what personalization aspects each group expect. 
To address these limitations, we propose a group-aware personalization framework, \method\ (\methodshort), that identifies context-specific variations in conversational preferences across user groups and then steers LLMs to address those preferences. 
Our approach consists of two steps: (1) Group-Aware Preference Extraction, where maximally divergent user-group preferences are extracted from real-world conversation logs and distilled into interpretable rubrics,
and (2) Tailored Response Generation, which leverages these rubrics through two methods: a) \texttt{Context-Tuned Inference} (\methodCT), that dynamically adjusts responses via context-dependent prompt instructions, and b) \texttt{Rubric-Finetuning Inference} (\methodFT), which uses the rubrics to generate contrastive synthetic data for personalization of group-specific models via alignment. 
Experiments demonstrate that our framework significantly improves alignment of the output with respect to user preferences and outperforms baseline methods, while maintaining robust performance on standard benchmarks.
\end{abstract}

\section{Introduction}
Large Language Models (LLMs) are pivotal in modern natural language processing (NLP), driving applications such as conversational agents, content generation, and automated reasoning~\citep{liu2024llmconversationalagentmemory, tian-etal-2024-large-language, mondal-etal-2024-presentations}. 
Despite their remarkable capabilities, LLMs often fall short in addressing the specialized needs of distinct user groups due to their \texttt{one-size-fits-all} training paradigm~\citep{lucy-etal-2024-one}. This approach predominantly relies on asking human or LLM judges to provide ratings (e.g. preferred and dispreferred labels) to alternative outputs for the same input query, in order to create {\em paired} preference data~\cite{ji2024aialignmentcomprehensivesurvey}. 
These approaches assume that human and AI annotators accurately reflect the preferences of the target user population. When models are aligned to this preference data, model outputs will be steered toward the most prevalent preferences of the {\em annotator} population, even when users express diverse preferences for the same task/query.   

Broad preference alignment like this can lead LLMs to produce suboptimal outputs for a {\em target} user base for two primary reasons. First, the distribution of preferences in the target population may differ from those expressed in the annotator population. 
Examples include domain-specific expertise (i.e., if annotators are generally non-experts, but the target users are experts)
and cultural norms (e.g., Japanese audiences may prefer narratives on family bonding, while U.S. audiences favor individualistic themes). 
Second, even across populations, preference differences may vary with respect to domain/task.
For instance, in education, experts may expect precise terminology and assume foundational knowledge, while novices may desire real-world analogies and step-by-step explanations. In programming, experts often prefer concise debugging strategies, whereas novices may seek explicit concept explanations with visual aids. 

\begin{figure*}
    \centering
    \fbox{\includegraphics[ 
    width=0.87\linewidth]{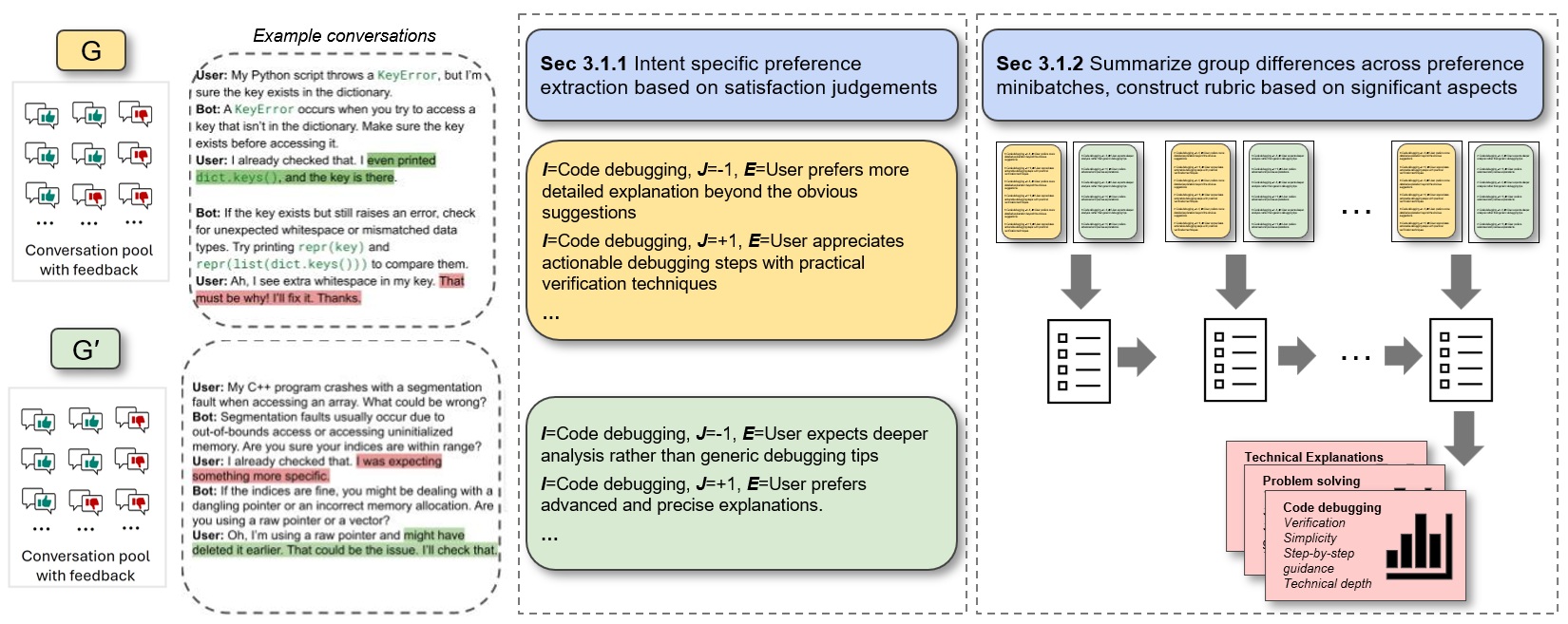}}
    \vspace{-2mm}
    \caption{Illustration of \methodshort rubric extraction (Sec.~\ref{subsec:extractpref}) showing group aware preference extraction across two groups ({\em Expert} v. {\em Novice}) with  conversations about Docker and .env file integration. First 
    individual preferences are extracted from conversations,  then the extracted preferences are grouped into minibatches, contrasted to extract salient differences, and summarized into intent-specific rubrics. 
    }
    \label{fig:rubric-extract}
    \vspace{-4mm}
\end{figure*}

\begin{figure}
    \centering
    \fbox{\includegraphics[
    width=0.95\linewidth]{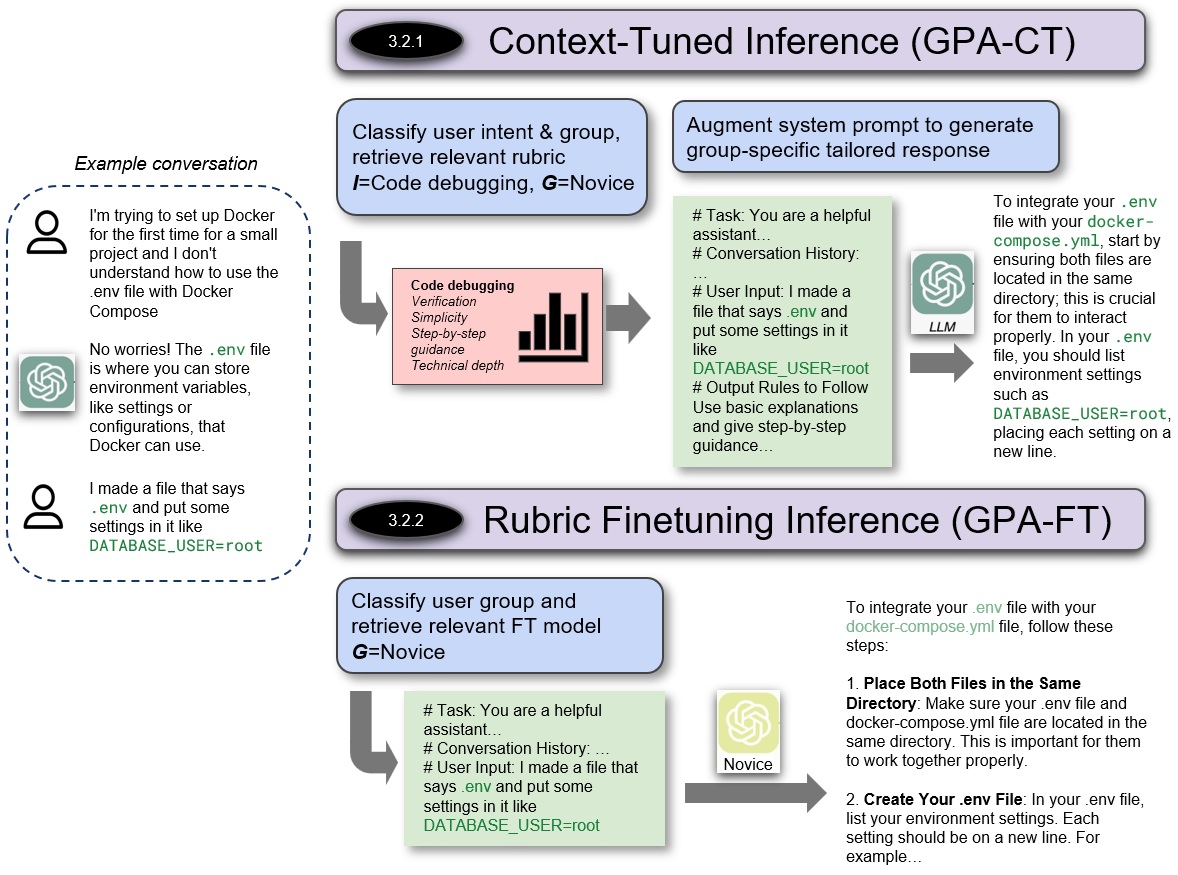}}
    \vspace{-2mm}
\caption{Illustration of \methodshort tailored response generation (Sec.~\ref{subsec:Alignment}) for a {\em Novice} user with a Code debugging intent using \methodCT and \methodFT.
    }
    \label{fig:gpa-inference}
    \vspace{-7mm}
\end{figure}

Existing methods for group-aware preference adaptation \cite{balepur2025boatdoesfloatimproving, li2024culturellmincorporatingculturaldifferences} focus on the first issue only and aim to improve response generation through the use of {\em personas}. These approaches either use  abstract descriptions of personas or auxiliary data reporting general group preferences (e.g., cultural norms) to generate synthetic preference data. However, these methods are limited by their use of external preference signals or internal LLM knowledge of likely preferences for a specified user segment. Specifically, they are unlikely to capture the full range of group-specific preferences expressed across various contexts, which can be directly observed via {\em in-situ} user interactions.  
As users prefer responses for varied
reasons \cite{Kirk2024TheBR}, models tuned on preference data should customize outputs to meet these group specific needs \cite{salemi-etal-2024-lamp, li2024culturellmincorporatingculturaldifferences} while also keeping in mind that intent further modulates user preferences.
This highlights the need for \textbf{group-aware contextual preference learning} to extract users’ diverse preferences for tailored response generation.

In this paper, we aim to address this issue by proposing a novel group-aware customization framework, \method\ (\methodshort), that automatically identifies context-specific variations in conversational preferences across user groups and steers LLMs to address those preferences.  Our approach consists of two components. \textbf{First}, we propose a method to extract salient group preference differences from real-world conversation logs. We then distill the output into interpretable rubrics that summarize intent-specific guidance for each group (see Fig.~\ref{fig:rubric-extract} for illustration).  Our method is generalizable across user groups and notably, if the groups do not express significant preference differences on specific intents, the returned rubric set will be empty,
hence signifying no need for customization. \textbf{Second}, we propose two methods to use these rubrics to tailor personalized responses for each group (see Fig.~\ref{fig:gpa-inference} for illustration): (1) \texttt{Context-Tuned Inference} (\methodCT) dynamically adjusts responses via context-dependent in-prompt augmentation, which is data-efficient 
and training-free, and (2) \texttt{Rubric-Finetuning Inference} (\methodFT) uses the learnt rubrics to generate  contrastive synthetic data to fine-tune separate models towards group-specific preferences. 

We evaluate \methodshort on two in-situ conversational datasets (WildChat~\cite{zhao2024wildchat1mchatgptinteraction} and Microsoft Copilot logs). Our experiments using using LLM persona-guided evaluation \cite{koutcheme2024opensourcelanguagemodels, dong-etal-2024-llm} show that models customized with \methodshort\ outperform all baseline methods, including static-preference, persona-guided and zero-shot models. In addition we demonstrate that \methodshort improves user satisfaction. 
Notably, alignment with \methodshort\ produces these improvements without compromising LLM’s core capabilities, as evidenced by robust performance on standard benchmarks such as MT-Bench \cite{zheng2023judgingllmasajudgemtbenchchatbot} and Arena-Hard \cite{li2024crowdsourceddatahighqualitybenchmarks}.

\section{Problem Definition and Notations}

We hypothesize that intent-driven user preferences can be automatically extracted from real-world conversation logs between human and AI agents, enabling more effective model alignment than traditional methods that do not incorporate direct user feedback.  
Consider a user group \( \gG \) that generates queries for a specific intent \( \gI \)\footnote{To simplify notation, we use {\em intent} to encode both {\em domain} (eg. education) and {\em task} (eg. summarization).}. The responses from the LLM, denoted as \( Y_{\gI} = \text{LLM}(X_{\gI}) \), receive user judgments \( \gJ_{\gG}(Y_{\gI}) \) in the form of thumb feedback or implicit textual feedback (eg. thanking the AI). When these preferences diverge from the general population’s judgments \( \gJ_P(Y_{\gI}) \), we hypothesize that aligning the model with group-specific signals will improve response relevance and user satisfaction. Note that if \( \gJ_{\gG}(Y_{\gI}) \approx \gJ_P(Y_{\gI}) \), alignment to \( \gJ_{\gG}(Y_{\gI}) \) will simply reinforce existing preferences in the general population without degrading performance.  
Unlike RLHF~\citep{ouyang2022traininglanguagemodelsfollow} and RLAIF~\citep{bai2022traininghelpfulharmlessassistant}, which optimize for majority preferences, our approach leverages in-situ user judgments to achieve fine-grained, group-specific alignment, that is of particular use when user needs deviate significantly from broader norms.

Let \( C = \{C_1, C_2, \dots, C_n\} \) represent a set of conversations from a collection of users, where each \( C_i \) is an individual conversation.
Let each conversation \( C_i \), consisting of \( t \) interaction turns of user-agent utterances, be represented as: $C_i = [U_1, A_1, \dots, U_t, A_t]$. 
Here, \( U_t \) refers to a user utterance and \( A_t \) refers to an AI agent response. The user-agent conversations \( C_i \) often consist of multiple turns, i.e., \( t \geq 1 \).
Each conversation $C_i$ is labeled with a predicted intent $\gI_i$ (see e.g., \citet{TnT-LLM}). Each conversation turn $U_t$ has been labeled with a user satisfaction judgment \( \gJ_i \in [-1, +1] \) using \citet{lin-etal-2024-interpretable}.   
Finally, we assume that each user $u$ is associated with one of two groups, ie. $u \in \gG$ or $u \in \gG'$. 
Note that in cases where contrasting group labels are unavailable, \methodshort\ can also be used by comparing a single group \( \gG \) against the overall population \( P \). We do not make any assumptions about \(| \gG |\) as long as there are sufficient interactions from users in \( \gG \) to extract preferences.

\section{Group Preference Alignment (GPA)}

Our \methodshort\ framework enables \textbf{context-aware, group-specific adaptation}, ensuring more precise and effective model alignment beyond more conventional preference optimization using auxiliary annotators.
The overall approach to align models with in-situ preferences involves two main steps: (i) Generating rubrics with group-aware preference extraction (Section~\ref{subsec:extractpref}), and (ii) Tailoring responses based on the extracted rubrics (Section~\ref{subsec:Alignment}). We discuss each in more detail below.

\subsection{Group-Aware Preference Extraction}
\label{subsec:extractpref}
\methodshort\ automatically identifies context-specific variations in conversational preferences across user groups \( \gG \) and \( \gG' \) and summarizes the divergent preferences into rubrics (Figure~\ref{fig:rubric-extract} and Algorithm \ref{algo:rubricextraction}). 
Specifically, given conversations regarding  specific intents \( \gI \) from users in \( \gG \) and \( \gG' \), with satisfaction judgments \( \gJ \) from user responses, the algorithm  uses the judgments to infer individual preferences \( \mathcal{E} \) that explain the user's positive or negative feedback (Section~\ref{subsubsection:grouping}). 
Next, the preferences are summarized into generalized preference aspects \( \gA \), capturing salient differences between two groups (Section~\ref{subsubsection:update}). 
The resulting group-specific rubrics serve as the foundation for group-aware customization.

\subsubsection{Extract Intent-Specific Preferences}
\label{subsubsection:grouping}

Algorithm \ref{algo:rubricextraction} (Appendix, Lines 1-13) show how we learn group-specific preference rubrics based on user conversations and their corresponding intent labels or context. The input includes a conversation set \(C\), user groups \(\gG\) and \(\gG'\), intent labels \(\gI\), user judgments \(\gJ\), a Likert scale threshold \(\ell\), and a minibatch size \(m\). The algorithm processes each conversation \(C_i\) consisting of \(t_i\) interaction turns. For each turn \(S_j\), the algorithm checks whether the turn is associated with  a user  judgment \(\mathcal{J}(S_j)\). If the turn expresses implicit satisfaction (SAT) or dissatisfaction (DSAT) (ie. \(abs(\mathcal{J}(S_j)) = 1\)), we use an LLM to infer individual preferences and generate an explanation (\(\gE_+\) or \(\gE_-\) for SAT and DSAT judgments respectively) [Prompts in Table~\ref{table:sat_InferUserExpectation} and Table~\ref{table:dsat_InferUserExpectation}, Appendix].
These preferences, \(\gE_+\) and \(\gE_-\), are then grouped by intent \(\gI_k\) for each user group \(\gG\) and \(\gG'\). 
At the end of this phase, the preferences are organized into sets by user groups, intents and satisfaction.

\subsubsection{Summarize Group Differences}
\label{subsubsection:update}
Algorithm \ref{algo:rubricextraction} (Appendix, Lines 14-30) describe how group-specific preferences are summarized into  intent-specific rubrics. 
For each intent \( \gI_k \in \gI \), the algorithm partitions the preferences of each user group ($\gG, \gG'$) into minibatches of size \( m \). 
The algorithm then iterates over pairs of minibatches, one from each group, and summarizes/updates the divergent aspects of their expressed preferences. Specifically, for a pair of minibatches (\( \mathcal{E}_{G,I_k}^a \) and  \( \mathcal{E}_{G',I_k}^b \)), the algorithm extracts a set of {\em aspects} \( \gA \) that summarize how a preference differs across the two groups (see Figure~\ref{fig:rubric-extract} for illustration). [Prompt in Table~\ref{table:computelikert}, Appendix].
The algorithm also estimates a divergence score \( r \) based on Likert scale to rate the significance of each aspect. 
If the divergence score \( r \) exceeds a threshold \( \ell \), indicating a significant difference in preferences between the two groups, the aspect \( \gA_{ab} \) is added to the rubric for the current intent \( \gR_{I_k} \).
Each iteration is provided the aspects from the previous round, so the algorithm can update/refine the aspects as it processes all the minibatches. The process continues until all significant divergent aspects have been identified and included in \( \gR_{\gI_k} \). Finally, the rubric for this intent is added to the global rubric list \( \gR \). 
The algorithm returns the full set of rubrics \( \gR \) 
and an interpretation of each rubric. These capture  
 the distinct preference patterns of the two user groups across intents. Fig.~\ref{fig:rubrics_significant} shows some example rubrics extracted from Microsoft Copilot conversations.

\subsection{Response Tailoring}
\label{subsec:Alignment}
After learning rubrics in section~\ref{subsec:extractpref}, we outline the following two methods to align LLMs based on these learnt rubrics/preferences. 
The first method, \methodCT, involves dynamically augmenting prompts to produce group-aware tailored responses. It dynamically adjusts the LLM prompts incorporating learnt rubrics from~\ref{subsec:extractpref} during inference, conditioned on intent and user group identified for each conversation (Section~\ref{subsubsection:contextuning}).
The second approach, \methodFT\ uses learnt rubrics to synthetically augment conversational data with paired responses that reflect group-specific preferences conditioned on specific intents.  This process produces a tailored set of preference data. We finetune a group-specific LLM using this enriched dataset to enhance their alignment to the targeted group (Section~\ref{sec:finetuning}).

\subsubsection{\methodCT: Dynamic Context-Tuning}
\label{subsubsection:contextuning}

Context-tuning with \methodCT\ is an adaptive process that infers the user's group and intent, retrieves the relevant rubrics for that intent, and then modifies the instructions sent to the LLM to generate the next output (Illustration in Fig.~\ref{fig:gpa-inference} top row; Algorithm~\ref{algo:context-tuning}, Appendix). 
Unlike finetuning, which  adjusts a model's weights based on a fixed training dataset, context-tuning allows for dynamic adjustments to the model's prompt based on real-time analysis of user intent and group membership. This means the model can adapt to the specific needs of different groups on-the-fly without requiring specialized group-specific models.
The advantage of \methodCT\ over finetuning includes the flexibility to adapt to user-specific needs without retraining and enhanced efficiency (as it avoids the extensive resources typically required for finetuning).

\begin{figure*}[!t]
    \centering
    \includegraphics[width=0.99\linewidth]{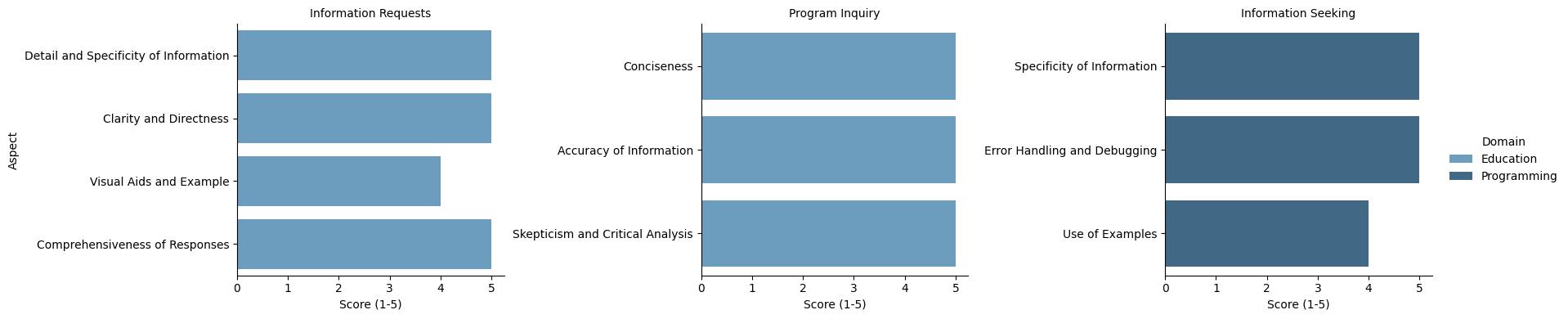}
    \vspace{-2mm}
    \caption{Rubrics/Aspects on which Experts and Novices Differ in the Education and Programming Domains as extracted from Microsoft Copilot logs on three tasks (Information Requests, Program Inquiry, Information Seeking) with the Likert-Scale Rating (1-5) on the x-axis and aspect/rubric names on the y-axis.}
    \label{fig:rubrics_significant}
    \vspace{-1mm}
\end{figure*}

\subsubsection{\methodFT: Rubric-Guided Contrastive Data Generation and  Fine-tuning}
\label{sec:finetuning}

Rather than merely fine-tuning LLMs with the training data comprising of preference signals from user groups (\( \gG \) and \( \gG' \)), we use our learned rubrics to generate more realistic contrastive pairs that vary according to observed preference dimensions in the extracted rubrics. \methodFT\ may be favored over \methodCT\ in situations where the LLM is less steerable with prompt-tuning (eg. smaller models) and/or when lower latency is desired for inference. 

Specifically, we finetune LLMs using synthetic training data generated with intent- and group-aware rubrics to reflect in-situ user preferences.
Algorithm \ref{algo:data-generation} (Appendix) describes our approach to rubric-guided data generation. It takes as input a conversation and augments the existing AI response with a paired response of opposing preference polarity, conditioned on the group-aware rubrics.
Consider a conversation $S_i$ up to the $j^{th}$ user utterance, with $A_j$ referring to the corresponding AI response, and $\gJ(S_i) \in \{+1, -1\}$ referring to the user satisfaction judgment for $A_j$. To generate a {\em contrastive augmented sample}, we modify the response as follows: If $\gJ(S_i) = +1$ (preferred response), we generate a \textit{dispreferred response} $A_{aug}$ by instructing the LLM to incorporate features from the opposing group’s rubric for that intent. Otherwise, if $J(S_i) = -1$ (dispreferred response), we generate a \textit{preferred response} $A_{aug}$ by instructing the LLM to align the output with the user's group rubric for that intent.
When applied to the full training data, the procedure produces an augmented dataset $\mathcal{D}_{aug}$ where each original instance is paired with a contrastive sample:
$T_{aug}$ = ($S_i$, $A_{aug}$, $-\gJ(S_i)$).
Note that $\gJ_{aug} = -\gJ(S_i)$ to faciliate contrastive preference learning.
Next, 
we train separate models for each user group. Given a prompt $S$ and responses $A_{+}$ (preferred) and $A_{-}$ (dispreferred), the likelihood of selecting the preferred response is modeled as:

\vspace{-6mm}
\begin{equation}
P_{\theta}(A_{+} | S) = \frac{e^{f_{\theta}(S, A_{+})}}{e^{f_{\theta}(S, A_{+})} + e^{f_{\theta}(S, A_{-})}}
\end{equation}

\noindent where $f_{\theta}(S, A)$ is a scoring function parameterized by $\theta$, representing the model’s preference alignment. The DPO objective is to maximize the log-likelihood of the chosen response:

\vspace{-6mm}
\begin{equation}
\mathcal{L}_{DPO} = \mathbb{E}_{(S, A_{+}, A_{-}) \sim \mathcal{D}_{aug}} \left[ \log P_{\theta}(A_{+} | S) \right]
\end{equation}

By optimizing $\mathcal{L}_{DPO}$, the model learns to \textbf{prefer responses aligned with group-specific rubrics} while discouraging responses reflecting dispreferred aspects.
We train two specialized DPO models ($P_{\theta_G}$ and $P_{\theta_{G'}}$) using contrastive samples from $\mathcal{D}_{aug}$ for each  
(see Algorithm~\ref{algo:rubricalign-train}, Appendix).
This results in:
\textbf{[1]} $P_{\theta_G}$, optimized to generate responses aligned with the preferences of user group $G$.
\textbf{[2]} $P_{\theta_{G'}}$, optimized for user group $G'$. Note that we experimented with KTO objective as well, but found DPO to be superior (Table~\ref{tab:choice_of_dpo_vs_otherFT}, Appendix).

See Fig.~\ref{fig:gpa-inference} (bottom row) for an illustration of how we apply the finetuned models for inference.  Algorithm~\ref{algo:rubricalign-infer} (Appendix) specifies details.

\begin{table*}[h]
\small
    \centering
    \begin{tabular}{lcccc}
        \toprule
        \textbf{Model} & \textbf{LLM Pref (W/L/T)} &  \textbf{LLM conf $\geq$ 75} & \textbf{LLM Pref (W/L/T)} &  \textbf{LLM conf $\geq$ 75} \\
        \midrule
                &  \multicolumn{2}{c}{\textbf{Intent=Programming/Group=Novice}} & \multicolumn{2}{c}{\textbf{Intent=Programming/Group=Expert}}  \\
        \midrule
        \methodCT\ vs Base & 65.82 / 25.00 / 9.18 & 67.53 / 32.47 & 57.10 / 42.04 / 0.86 & 57.46 / 42.54 \\
        \methodCT\ vs Persona & 60.44 / 31.96 / 7.60 & 73.97 / 26.3 & 61.10 / 38.30 / 0.6 & 61.91 / 38.09\\
        \methodCT\ vs Static & 56.43 / 37.43 / 6.14 & 80.00 / 20.00 & 57.38 / 41.47 / 1.6 & 59.05 / 40.95\\        
        \midrule
        \methodFT\ vs Base & 71.29 / 25.87 / 2.84 & 68.05 / 31.95 & 53.17/ 40.62 / 5.56 & 56.15 / 43.84\\
        \methodFT\ vs Persona & 70.98 / 27.76 / 1.26 & 68.84 / 31.16 & 58.80 / 40.62 / 5.0 & 59.62 / 40.37\\
        \methodFT\ vs Static & 66.88 / 32.18 / 0.95 & 60.64 / 39.36 & 59.65 / 39.77 / 0.56 & 57.72 / 42.27\\
        \midrule
        \methodFT\ vs \methodCT\ & 63.09 / 36.59 / 0.32 & 57.59 / 42.41 & 53.12 / 38.35 / 0.28 & 58.99 / 41.00\\
        \midrule
                &  \multicolumn{2}{c}{\textbf{Intent=Writing/Group=USA}} & \multicolumn{2}{c}{\textbf{Intent=Writing/Group=China}}  \\
        \midrule
        \methodCT\ vs Base & 45.5 / 53.5 / 1.0 & 54.1 / 45.9 & 58.5 / 23.9 / 17.6 & 88.57 / 11.42 \\
        \methodCT\ vs Persona & 55.5 / 42.5 / 2.0 & 59.5 / 40.5 & 53.6 / 28.73 / 17.60 & 60.0 / 40.00\\
        \methodCT\ vs Static & 67.02 / 31.00 / 1.98 & 67.10 / 32.90 & 52.11 / 32.3 / 15.59 & 68.57 / 31.43\\        
        \midrule
        \methodFT\ vs Base & 55 / 26.5 / 18.5 & 62.2 / 37.8 & 55.22 / 20.84 / 23.94 & 60.95 / 39.04 \\
        \methodFT\ vs Persona & 77 / 21.5 / 1.5 & 82.4 / 17.5 & 35.21 / 40.84 / 23.95 & 32.38 / 67.62 \\
        \methodFT\ vs Static & 85 / 14.5 / 0.5 & 88.5 / 11.5 & 28.16 / 54.92 / 16.92 & 47.61 / 52.39\\
        \midrule
        \methodFT\ vs \methodCT\ & 85.5 / 14 / 0.5 & 71.4 / 28.6 & 39.43 / 40.84 / 19.73 & 40.95 / 59.05\\
        \bottomrule
    \end{tabular}
    \caption{WR Evaluation of \methodCT and \methodFT on Wildchat Creative Writing  across countries, and Microsoft Copilot Programming across experts/novices. W/L/T=win/lose/tie rates
    and 
    LLM confidence~\cite{dong-etal-2024-llm}.}
    \label{tab:full_results_llama}
\end{table*}

\section{Experimental Setup}

We evaluate 
$\methodshort$ 
using real-world conversational logs from Microsoft Copilot and Wildchat (\citet{zhao2024wildchat1mchatgptinteraction}) data.
For the intent category {\em programming and software}, we use 8000 Copilot conversations  and 8200 WildChat conversations. 
We group conversations into {\em expert} (i.e., $\gG$; Copilot: 2200, WildChat: 6000) and {\em novice} (i.e., $\gG'$; Copilot: 5800, WildChat: 2000) groups by using an auxiliary {\em expertise} classifier \footnote{Prompt in Table~\ref{table:expertise}, Appendix. 
We  manually inspected ~100 random conversations and found that the classification was reliable ($\kappa=0.88$ agreement computed between the first author and GPT-4o).},  (Copilot: 5800 novice,  2200 expert; WildChat: 6000 expert, 2000 novice).
Next, we consider the intent category {\em Creative writing and editing} in WildChat and form user groups based on metadata, specifically {\em location}, partitioning users into 
USA (8000 conversations) and 
China (800 conversations) (Full Dataset Statistics in Appendix).
We partitioned the above datasets into 90:10 train:test split to ensure no training signal leakage. 
Next, we use predicted SAT/DSAT  judgments to learn divergent preferences on the training data. 
Following \citet{lin-etal-2024-interpretable} and the taxonomy proposed by \citet{shi2024wildfeedbackaligningllmsinsitu}, we used GPT-4o to classify bot responses resulting in a subsequent user SAT, DSAT, or Neither judgment.
We finetune using synthetic data constructed from the full training set.

\paragraph{Models and $\methodshort$ Baselines.} 
For rubric extraction, we use GPT-4o and for  tailored response generation (\methodCT and \methodFT), we use two base LLMs ($M$): \texttt{gemma-2-9b-it}\footnote{{\tiny\url{https://huggingface.co/google/gemma-2-9b-it}}} \cite{gemmateam2024gemma2improvingopen} and \texttt{Meta-Llama-3-8B}\footnote{{\tiny\url{https://huggingface.co/meta-llama/Meta-Llama-3-8B}}} \cite{grattafiori2024llama3herdmodels}.
We compare \methodCT and \methodFT  against several baselines: a) \textbf{Zero-shot (Base)} responses,
b) \textbf{Persona-Aware (Persona-{G})}: which augments the input prompt with group-aware persona (G) information to mimic responses from specific user-groups through role-playing behavior, (Prompt in Table~\ref{table:persona-aware}, Appendix) 
c) \textbf{Persona-Criteria-Aware (Static-{G})}: which uses $M$ to first  generate preference criteria for $G$ and $G'$, and then append the generated criteria to the prompt (Prompt in Table~\ref{table:static}, Appendix),
d) \textbf{KTO (KTO-{G})}: which fine-tunes an LLM with group-specific SAT and DSAT samples to tailor towards each group using KTO \cite{ethayarajh2024ktomodelalignmentprospect}, 
e) \textbf{KTO-Augmented (KTO-{G$^{+}$})}: which also uses KTO to finetune an LLM on the group-specific SAT and DSAT samples, but this time augmented with the contrastive pairs generated by our rubrics.

\paragraph{Evaluation Metrics.}

We evaluate responses across three key dimensions:
\textbf{1) Customization to Group Preferences}: We assess alignment with group-specific preferences using Win-Tie-Lose (WTR) rates computed via GPT-4o-as-a-Judge with Persona-Role Playing \cite{dong-etal-2024-llm} (Prompt in Table~\ref{table:PersonaEvaluation_withoutEC}, Appendix).
We also report WTR results for LLM judgments  with confidence estimations \cite{dong-etal-2024-llm}, at a confidence threshold correlated with human judgment (Prompt in Table~\ref{table:PersonaEvaluation_withEC}, Appendix). To mitigate positional bias, we average win rates by swapping response positions.
\textbf{2) Oracle-Guided Satisfaction Estimation}: We identify responses that significantly deviate from DSAT-classified reference responses to minimize negative follow-up feedback. This evaluation measures whether our methods generate fewer dissatisfactory signals than baselines, focusing solely on response differences without considering user personas (Prompt in Table~\ref{table:ind_dsat_evaluation}, Appendix). Success is determined by the number of times our responses outperform baselines in WTR comparisons.
\textbf{3) Quality Evaluation on Standard Benchmarks}: We assess the generalization of \methodCT and \methodFT using two open-ended instruction-following benchmarks: MT-Bench \cite{zheng2023judgingllmasajudgemtbenchchatbot} and Arena-Hard \cite{li2024crowdsourceddatahighqualitybenchmarks}. This ensures that our models maintain strong performance in general instruction-following tasks despite group-aware alignment. We follow each benchmark’s evaluation protocol, reporting Win Rate (WR) for Arena-Hard (with GPT-4o as the judge) and the average MT-Bench score using default inference strategies.

\section{Main Results and Findings}
\label{sec_results}

\paragraph{$\methodFT$ excels when ample finetuning data is available, while $\methodCT$ remains robust in lower-data settings.}
Table~\ref{tab:full_results_llama} reports the evaluation of \methodCT and \methodFT on Wildchat Creative Writing  across countries and Copilot Programming across experts/novices for Llama base model.
$\methodFT$ using Llama consistently outperforms all baselines when sufficient finetuning data is available, particularly in the Novice and US groups. In the Novice category, $\methodFT$ achieves a 71.29\% Win Rate (WR) vs. Base, compared to 65.82\% for $\methodCT$, showing that finetuning helps models better adapt to novice preferences. A similar trend is observed in the US Writing group, where $\methodFT$ achieves an 85\% WR vs. Static, outperforming $\methodCT$ at 67.02\% WR. 
Conversely, $\methodCT$ is more effective when data is scarce (also evident in Figure~\ref{fig:lc}, Appendix), as seen in China and Expert groups. Against the Base model, $\methodFT$ achieves only a 55.22\% WR in China, while GPA-CT performs slightly better at 58.5\% WR, indicating that in-context adaptation is more useful in this setting. 
These observations also hold on the Wildchat Dataset where we observe a drop in the WTR of \methodFT compared with \methodCT (Table~\ref{tab:wildchat_llama}, Appendix), since the Novice Groups have correspondingly fewer samples.

\paragraph{Base models such as Llama/Gemma are tuned more towards US Preferences and Expert Groups compared to China/Novice Groups.}
Table~\ref{tab:full_results_llama} shows that for Expert groups, Base is the hardest baseline to beat, reinforcing the fact that pretrained models are already aligned with expert preferences.
For example, $\methodFT$ vs. Base in Expert groups achieves only a 53.12\% WR, compared to 71.29\% WR for Novice users, indicating that Base already reflects expert-style responses well. The same trend appears in US-based writing conversations, where $\methodCT$ and $\methodFT$ struggle more against the Base model than against Persona or Static baselines, proving that pretraining biases favor Western-aligned outputs. Similar observations also hold true when Gemma is used as the base model (Table~\ref{tab:full_gemma_results_bing}, Appendix).

\paragraph{\methodCT improves overall satisfaction compared to baselines.} Results in Table~\ref{tab:eval3} (computed on the DSAT Signals in the Wildchat Programming) using Llama show that \methodCT  consistently wins against all setups, achieving the highest win rate against Static (76.70\%), followed by Base (69.61\%) and Persona (65.69\%).
This suggests that \methodCT generates responses that better align with user expectations and reduces dissatisfaction signals when compared with other baselines.

\begin{table}[!t]
\small
\centering
\label{tab:win_Lose_tie_percentage}
\begin{tabular}{lccc}
\toprule
\textbf{Setup} & \textbf{Win (\%)} & \textbf{Lose (\%)} & \textbf{Tie (\%)} \\
\midrule
\methodCT\ vs Base   & 69.61\% & 29.41\% & 0.98\% \\
\methodCT\ vs Persona & 65.69\% & 33.33\% & 0.98\% \\
\methodCT\ vs Static  & 76.70\% & 21.36\% & 1.94\% \\
\bottomrule
\end{tabular}
\caption{WTR against the baselines (Win determines the number of times \methodCT is chosen over others)  on Wildchat Programming when compared against reference DSAT Evaluation using Llama.}
\label{tab:eval3}
\end{table}

\paragraph{\methodFT\ does not compromise model performance on other benchmarks.}
We evaluate our LLama-based $\methodFT$ models on MT-Bench and Arena-Hard to assess if their instruction-following performance degrades on standard benchmarks as done by~\citet{shi2024wildfeedbackaligningllmsinsitu}. 
Table~\ref{tab:arena-hard} confirms that $\methodFT$ does not compromise performance on the instruction-following benchmarks. The MT-Bench scores for $\methodFT$ models remain close to the Base model (8.320), with Novice $\methodFT$ (8.334) even slightly outperforming it.  
Similarly, Arena-Hard results show a positive win-loss delta ($\Delta$) across all groups, with China $\methodFT$ (+10.15\%) and Novice $\methodFT$ (+10.01\%) achieving the highest gains. 
Even in Expert (+5.01\%) and US (+3.76\%) categories, $\methodFT$ maintains competitive performance. Overall these findings reinforce the fact that group preference alignment via fine-tuning does not lead to overfitting or loss of generalization.

\begin{table}[!t]
\footnotesize
    \centering
    \begin{tabular}{lcccrc}
        \hline
        \textbf{Model} & \textbf{WR} & \textbf{LR} & \textbf{TR} & \textbf{$\Delta$ (\%)} &
        \textbf{MT-B}\\
        \hline
        Base & & & & & 8.32 \\
        Novice \methodFT & 49.11 & 39.10 & 8.62 & +10.01 & 8.33\\
        Expert \methodFT & 47.89 & 42.88 & 6.41 & +5.01 & 8.21 \\
        US \methodFT & 47.56 & 43.80 & 8.64  & +3.76 & 8.26 \\
        China \methodFT & 48.49 & 38.34 &  9.02 &   + 10.15 & 8.30 \\
        \hline
    \end{tabular}
    \caption{Comparison against LLama Base on Arena-Hard Benchmark (Win/Lose/Tie, and Win-Lose $\Delta$) and evaluation on MT-Bench (MT-B).}
    \label{tab:arena-hard}
\end{table}

\section{Further Analysis}
\label{sec:further_analysis}

\paragraph{User preferences vary across cultures and domains.}
Table~\ref{tab:rubric_culture} (Appendix) listed extracted rubrics for Creative Writing and Editing across countries and tasks. It illustrates how cultural background significantly shapes user preferences, even for the same intent. US users prefer personal engagement and detailed narratives, while Chinese users favor clarity and structured summaries in writing and creative content. 
Similarly, when expertise remains constant but the domain shifts from Education to Programming, experts prioritize different aspects—educators value comprehensiveness and critical analysis, whereas programmers focus on specificity, error handling, and debugging (Figure~\ref{fig:rubrics_significant}).

\begin{figure}[!t]
    \centering
    \includegraphics[width=0.75\linewidth]{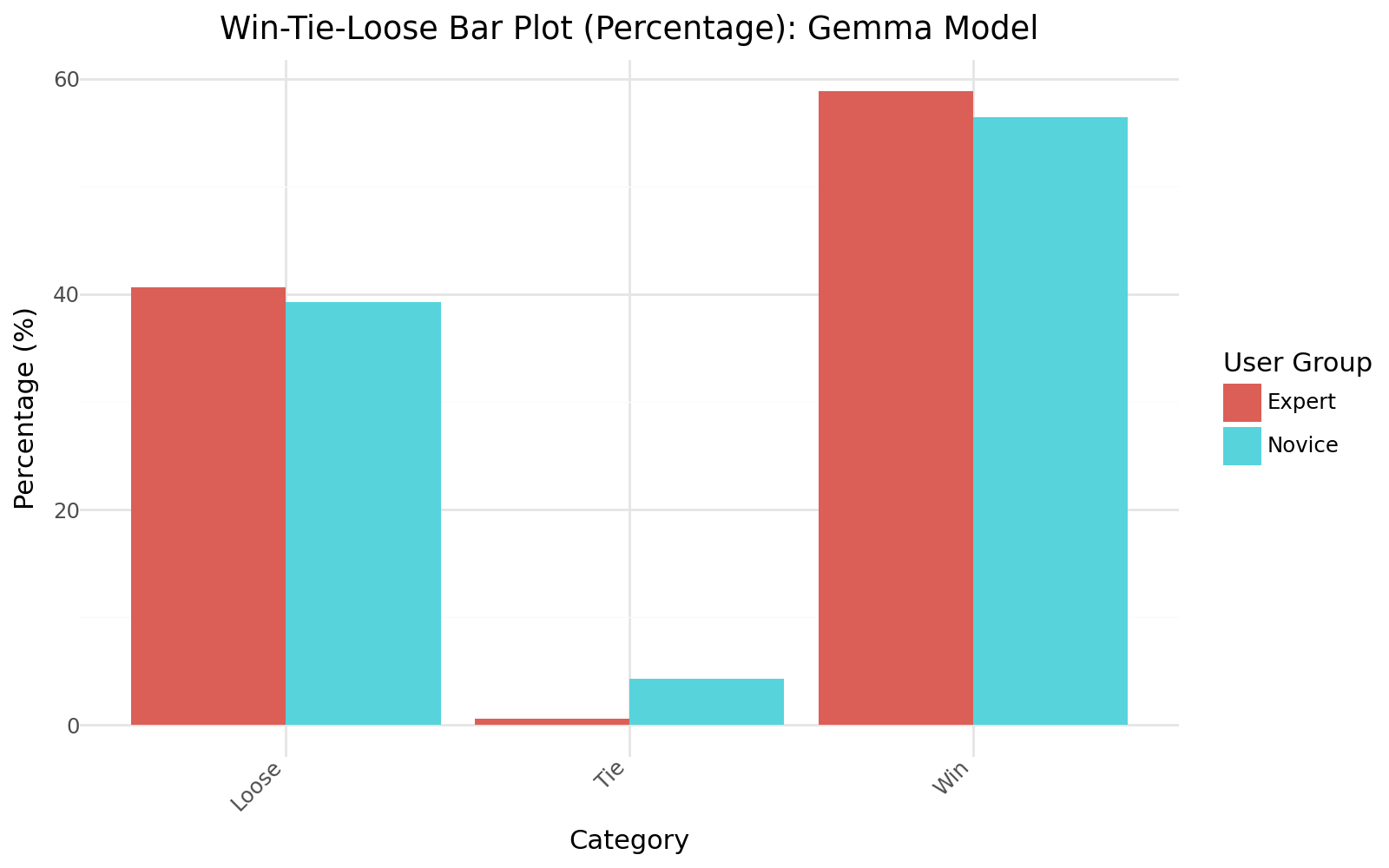}
    \vspace{-2mm}
    \caption{Bar plot evaluating Gemma 
    outputs from intent-aware rubric creation vs intent-unaware rubric creation using $\methodCT$. Results shows that intent heavily impacts performance when \methodshort  approach is used to personalize responses on Microsoft Copilot test set.}
    \label{fig:intent-aware}
    %\vspace{-1ex}
\end{figure}

\paragraph{Intent-Specific Rubrics are important for better group-preference alignment compared to generic ones.}
To investigate the impact of intents in preference learning, we extracted rubrics from the Microsoft Copilot expertise groups in two ways: without considering intent and with intent-awareness. 
These rubrics were then used for context-tuning on a held-out test set, followed by WTR evaluation using Persona-based evaluation. 
The results in Figure~\ref{fig:intent-aware} show a notable drop in WR when intent was not used , demonstrating that intent-aware rubric extraction leads to more personalized, contextually aligned responses.

\begin{figure}[!t]
    \centering
    \includegraphics[width=0.99\linewidth]{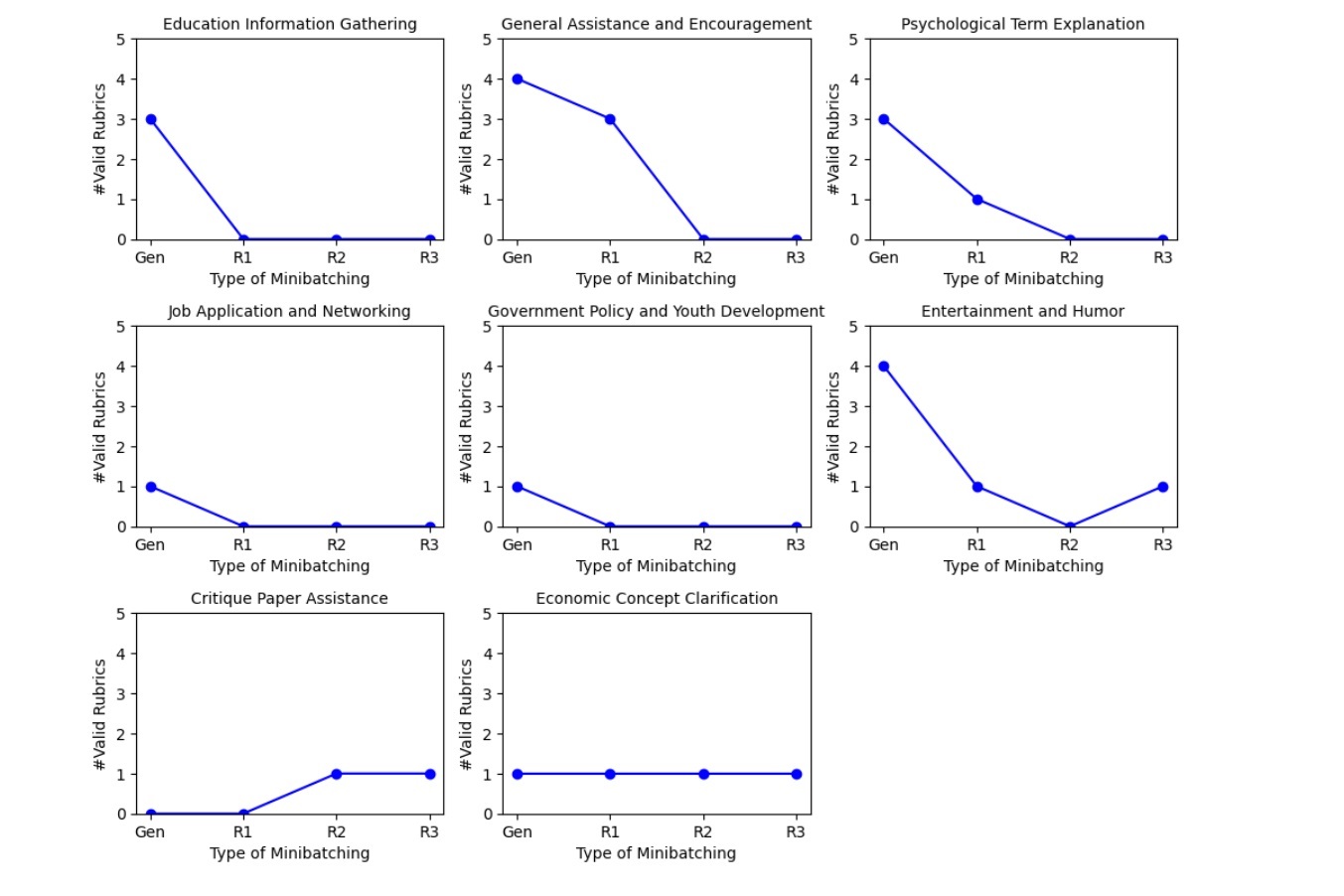}
    \vspace{-4mm}
    \caption{Illustrates how random shuffling of expertise labels impacts rubric generation. It reveals that, for most intents, shuffling results in the extraction of predominantly invalid rubric items, ultimately reducing the overall quality and number of valid rubric extractions.}
    \label{fig:robustness}
    \vspace{-4mm}
\end{figure}

\paragraph{Preference Rubrics degrade significantly when expertise labels are randomly flipped, indicating robustness.}
We assess the robustness of our extracted rubrics by randomly flipping expertise labels and extracting out the preference rubrics using \methodshort. Figure~\ref{fig:robustness} highlights the impact of random shuffling of expertise labels on rubric generation across various intents, where the validity of generated rubrics is determined by a self-correcting evaluation prompt from \methodshort. The results demonstrate that valid rubric generation is most successful under the original generation strategy (Gen) with correctly aligned expertise labels. However, as expertise labels are randomly shuffled (R1, R2, R3), the number of valid rubrics decreases significantly, often to zero, which confirm the robustness of our extracted rubrics.

\section{Related Work}

Customization of user interactions to better serve both individual and group preferences has a long history of research in a range of fields that leverage language technology. These include recommender systems~\cite{cho2002personalized,zhou2012state}, search and information retrieval~\cite{teevan2005personalizing,tabrizi2018person}, education~\cite{mchugh2020uncovering,domenichini2024llms}, and healthcare~\cite{wang2020using,li2024innovation}.

Meanwhile, LLMs are trained in a \texttt{one-size-fits-all} paradigm~\cite{lucy-etal-2024-one} where large-scale ratings from auxiliary human annotators or LLMs in a paired preference setup is used to teach models to generate preferred responses. This can make them difficult to customize. Nevertheless, recent work~\cite{zeng2023diverse,sorensen2024roadmap} has begun to advocate for the need for LLMs to serve more diverse preferences through pluralistic alignment.
Much of the work in LLM customization has focused on personalizing systems to the individual~\cite{Kirk2024TheBR,salemi-etal-2024-lamp}. These have used a variety of different approaches, including retrieval-augmented generation~\cite{salemi2024optimization}, memory~\cite{zhang2023memory}, parameter-efficient fine-tuning~\cite{tan2024democratizing}, and reinforcement learning~\cite{poddar2024personalizing}. Personalized LLM systems have also been applied to  diverse applications, such as contextual query suggestion~\cite{baek2024knowledge} and document creation~\cite{mondal-etal-2024-presentations}.

Recently some attempts have been made at modeling a large number of individual characteristics at scale, such as with a thousand preferences~\cite{lee2025aligning} or a million personas~\cite{ge2024scalingsyntheticdatacreation}. However, the focus on modeling \emph{group} preferences has been limited to a few recent research efforts~\cite{feng2024modular,zhao2023group,ramesh2024group}. Crucially, none of these methods leverage real-world conversational data at scale to {\em learn} these group preferences.
While some recent work has begun to incorporate feedback from in-situ user-AI interactions in order to improve models~\cite{shi2024wildfeedbackaligningllmsinsitu,li2024personalized}, their focus has been different from modeling group preferences. Thus, to the best of our knowledge, our paper is the first attempt at using large-scale satisfaction signals from human-AI conversation logs to customize LLM responses with group preference alignment (More in Appendix~\ref{sec:Relwork}).

\section{Conclusion} 

In this work we address a critical gap in group-aware personalization of LLMs by developing our Group Preference Alignment (GPA) framework. This framework identifies and incorporates diverse conversational preferences of distinct user groups  via a two-step process of Group-Aware Preference Extraction and Tailored Response Generation. Our experiments demonstrate that GPA significantly enhances the alignment of LLM outputs with group-specific preferences. GPA outperforms baseline methods with respect to preferences while maintaining robust performance on the standard information-following benchmarks. This work paves the way for developing more personalized and contextually aware LLMs using {\em in-situ} interactions using interpretable rubrics, which will ultimately improve user satisfaction and engagement. Due to increased transparency, this approach can be scaled up in legal/healthcare and other such high-stake domains.

\section*{Limitations}
While Group Preference Alignment (GPA) framework significantly improves user-group-specific response alignment, it has a few limitations: \\
\textbf{a) Dependence on Predefined User Groups}: The effectiveness of GPA relies on the availability of well-defined user groups with sufficient interaction data. In cases where user preferences are highly individualized or overlap across groups, extracting meaningful rubrics becomes challenging.\\
\textbf{b) Scalability of Rubric Extraction}: The framework extracts group-specific rubrics from conversation logs, which can be computationally expensive for large datasets. Additionally, the process assumes that conversational preferences remain stable within each group, which may not always be the case.\\
\textbf{c) Contextual Drift Over Time}: User preferences evolve, especially in dynamic domains like technology and education. The extracted rubrics may become outdated, requiring periodic updates to maintain alignment with current user expectations.\\
\textbf{d) Applicability Across Domains}: While GPA is tested on education, programming, and writing domains, its generalization to other highly specialized fields (e.g., legal or medical domains) remains unexplored. Future work should assess its adaptability to such contexts.
\bibliography{custom}
\appendix

\section{Appendix}
\subsection{Additional Related Work}
\label{sec:Relwork}
In this paper, we use LLMs as evaluators to measure the quality of system generations. Despite prior work pointing to some pitfalls with this approach, such as bias~\cite{koo2023benchmarking} and preferential scoring~\cite{liu2023llms}, using LLMs with judicious prompting for evaluation of language and information systems has become common practice~\cite{zheng2023judgingllmasajudgemtbenchchatbot,koutcheme2024opensourcelanguagemodels}. Recent efforts have applied the LLM-as-a-judge paradigm to evaluating a variety of applications such as translation~\cite{kocmi2023large} and summarization~\cite{jain2023multi}; notably these also include personalization~\cite{dong-etal-2024-llm}.

\subsection{Algorithm Pseudocode}
\label{sec:appendix:alg}

This appendix summarizes the pseudocode for \methodCT and \methodFT methods. Algorithm~\ref{algo:rubricextraction} depicts the procedure for extracting the group-aware preference rubric which is used in both methods. Inference for  \methodCT is next summarized in Algorithm~\ref{algo:context-tuning}. For \methodFT, we next describe the training procedure in Algorithm~\ref{algo:rubricalign-train}, and the pseudocode for generating the augmentented training examples for finetuning using the rubric is shown in Algorithm~\ref{algo:data-generation}. Finally, 
Algorithm ~\ref{algo:rubricalign-infer} applies the fine tuned models for \methodFT inference. We simply need to look up the appropriate group-aware model to use for generation.

\begin{algorithm*}[!t]
\small
\caption{Group-Aware Preference Extraction}
\begin{algorithmic}[1]
\Require Conversation set \( \mathbf{C} \); User groups \( \gG \) and \( \gG' \); Intent labels \( \mathbf{\gI} \); User judgments \( \mathbf{\gJ} \)
\Require Likert scale threshold $\ell$; Minibatch size \( m \) 
\Ensure Rubric \( \gR \)

\State \textbf{Step 1: Preference  Extraction}
\State \( \gE_+  = []; \gE_-  = [] \)
\For{each conversation $ C_i \in \mathbf{C}$ with \( t_i \) turns}
    \For{$j=[1..t_i]$}
        \State \( S_{j} = [U_1, A_1, \dots, U_j]_{C_i} \)
        \State \# If \( t_j \) contains judgment, extract preference
        \If{ $\gJ(S_{j})$ == +1  }
            \State  $\gE_+  = \gE_+\cup$ \{ \texttt{LLM.InferUserPreference} $(S_{j}, \gJ(S_{j}))$\} 
        \EndIf
        \If{$\gJ(S_{j})$ == -1}
            \State $ \gE_- = \gE_- \cup $ \{ \texttt{LLM.InferUserPreference} $(S_{j}, \gJ(S_{j}))$\}
        \EndIf
    \EndFor
    \EndFor
    \State \# Group prefs \( \gE_+ \) and \( \gE_- \) by intent \( I_k \) for each group
    \State \( \gE_{\gG, \gI_k} = \{\gE_+ \mid C_i \in \gG, C_i \text{ matches } \gI_k\} \cup \{\gE_- \mid C_i \in \gG, C_i \text{ matches } \gI_k\} \)
    \State \( \gE_{\gG', \gI_k} = \{\gE_+ \mid C_i \in \gG', C_i \text{ matches } \gI_k\} \cup \{\gE_- \mid C_i \in \gG', C_i \text{ matches } \gI_k\} \)

\State \textbf{Step 2: Aspect-Based Rubric Construction}
\State \# Initialize an empty list of rubric items 
\State \( \gR = [] \)
\For{each intent \( \gI_k \in \gI \) }
\State \# Uniformly partition each  explanation set into minibatches 
\State \( \gE_{\gG, \gI_k} = \{\gE_{\gG, \gI_k}^1, \dots, \gE_{\gG, \gI_k}^{n_1}\} \) s.t. $\forall a \; |\gE_{\gG, \gI_k}^a|=m$
\State 
\( \gE_{\gG', \gI_k} = \{\gE_{\gG', \gI_k}^1,  \dots, \gE_{\gG', \gI_k}^{n_2}\} \) s.t. $\forall b \; |\gE_{\gG', \gI_k}^b|=m$
\State \( \gA_{\gI_k} = [] \); $r_{\gI_k} = \{\}$

\For{each pair of minibatches \( (\gE_{\gG, \gI_k}^a, \gE_{\gG', \gI_k}^b) \)}
    \State \# Extract/update divergent aspects $\gA$ 
    \State \# Score group divergence on Likert scale $r$
    \State $[\gA_{ab},r_{ab}]$ = \texttt{LLM.ExtractAspectsAndLikert} \( (\gE_{\gG, \gI_k}^a, \gE_{\gG', \gI_k}^b, \gA_{\gI_k}) \)
    \State $\gA_{\gI_k} = \gA_{ab}$; $r_{\gI_k}[\gA_{ab}]=r_{ab}$
\EndFor
\State $\gR_{\gI_k}=[]$
\For{each aspect \( \gA_k \in \gA_{\gI_k} \)}
\If{\( r_{\gI_k}[\gA_{k}] > \ell \)}
    \State \( \gR_{\gI_k} \gets \gR_{\gI_k} \cup \{\gA_{k}\} \)
\EndIf
\EndFor

\State  \( \gR \gets \gR \cup \{\gR_{\gI_k}\} \)
\EndFor
\State \Return \( \gR \)

\end{algorithmic}
\label{algo:rubricextraction}
\end{algorithm*}

\begin{algorithm*}[!t]
\small
\caption{\methodCT: Inference}
\begin{algorithmic}[1]
\Require Partial conversation $S_{i} = [U_1, A_1, \dots, U_j]$ up to $j^{th}$ user utterance
\Require Rubric \( \gR \) 
\Ensure LLM answer $A_j$

\State \textbf{Step 1: Classify user group and intent}
\State $\gI_i =$ \texttt{Intent}($S_{i}$)
\State $\gG_i =$ \texttt{Group}($S_{i}$)

\State \textbf{Step 2: Retrieve Rubric and Augment Prompt}
\State $\gR_i = \gR_{\gI_i}$ 
\State $A_j $ = \texttt{LLM.ModifyPromptWithRubrics}(\( S_i, \gG_i, \gR_i \))
\State \Return \( A_j \)

\end{algorithmic}
\label{algo:context-tuning}
\end{algorithm*}

\begin{algorithm*}[!t]
\small
\caption{\methodFT\ Training}
\begin{algorithmic}[1]
\Require Conversation set \( \mathbf{C} \); User groups \( \gG \) and \( \gG' \); Intent labels \( \mathbf{\gI} \); User judgments \( \mathbf{\gJ} \); Rubrics \( \gR \)
\Require $Model_{Base}$ is the base LLM model 
\Ensure $Model_{FT}$ is fine-tuned model dictionary per group

\State \textbf{Step 1: Generate Synthetic Data} \For{each group $ {\gG},\gG'$}
    \State $T_{aug,\gG}$ = []
\EndFor
\For{each conversation $ C_i \in \mathbf{C}$}
    \State $\gI_i =$ \texttt{Intent}($C_{i}$)
    \State $\gG_i =$ \texttt{Group}($C_{i}$)
    \State $T_{aug,\gG_i}$ = $T_{aug,\gG_i} \cup \{C_{i}$\} 
    \For{$j=[1..t_i]$}
        \State \( S_{j} = [U_1, A_1, \dots, U_j]_{C_i} \)
        \State $S_{j,aug}$ = \texttt{RubricGuidedDataGeneration}($S_{j}$, $\gI_i$,$\gG_i$, \texttt{$\gR$}[$\gI_{i}$])
        \State $T_{aug,\gG_i}$ = $T_{aug,\gG_i} \cup \{S_{j,aug}$\}
    \EndFor
\EndFor
\State \textbf{Step 2: FineTune LLM for each group}

\State $Model_{FT}$ = \{\}
\For{each group $ {\gG} ,\gG'$}
    \State $Model_{FT}[\gG_i]$ = \texttt{FineTuneLlm}($Model_{Base}$, $T_{aug,\gG_i}$, $\gJ$)
\EndFor
\end{algorithmic}
\label{algo:rubricalign-train}
\end{algorithm*}

\begin{algorithm*}[!t]
\small
\caption{Rubric-Guided Data Generation}
\begin{algorithmic}[1]
\Require Training example $T = [S_i,A_j,\gJ(S_i)]$, where 
 $S_{i} = [U_1, A_1, \dots, U_j]$ is a conversation up to $j^{th}$ user utterance, $A_j$ is the AI response, and $\gJ(S_i)$ is the user's preference judgement for $A_j$ 
\Require Intent $\gI_i$, Group $\gG_i$, Rubric \( \gR_{\gI_i} \) 
\Ensure Augmented training data $T_{aug}$

\State \# Generate Augmented Training Example with Rubric
\If{ $\gJ(S_i)$ == +1  }
    \State \# Output is preferred by user, modify to include dispreferred group aspects 
    \State $A_{aug}$ = \texttt{LLM.ModifyPromptWithRubrics}(\( S_i, \gG', \gR_i \))
    \State $T_{aug} = [S_i,A_{aug},-1]$
\EndIf
\If{$\gJ(S_i)$ == -1}
    \State \# Output is dispreferred by user, modify to include preferred group aspects
    \State $A_{aug}$ = \texttt{LLM.ModifyPromptWithRubrics}(\( S_i, \gG_i, \gR_i \))
    \State $T_{aug} = [S_i,A_{aug},+1]$
\EndIf
\State \Return \( T_{aug} \)
\end{algorithmic}
\label{algo:data-generation}
\end{algorithm*}

\begin{algorithm*}[!t]
\small
\caption{\methodFT: Inference}
\begin{algorithmic}[1]
\Require Partial conversation $S_{i} = [U_1, A_1, \dots, U_j]$ up to $j^{th}$ user utterance
\Require Per-group, fine-tuned model dictionary \( Model_{FT} \) 
\Ensure LLM answer $A_j$

\State \textbf{Step 1: Classify user group}
\State $\gG_i =$ \texttt{Group}($S_{i}$)

\State \textbf{Step 2: Retrieve Group-Aware Model and generate response}    
\State $Model_{FT}$ = $Model_{FT}[\gG_i]$\State $A_j $ = $Model_{FT,\gD_i}(S_{i})$
\State \Return \( A_j \)

\end{algorithmic}
\label{algo:rubricalign-infer}
\end{algorithm*}

\subsection{Additional Experimental Results}

Rubrics extracted for US/China groups for Creative Writing and Editing can be found in Table~\ref{tab:rubric_culture}.

\begin{table*}[]
    \centering
    \renewcommand{\arraystretch}{1.1} % Adjust row spacing
    \setlength{\tabcolsep}{3pt} % Adjust column spacing
    \small % Reduce font size to fit within a page

    \begin{adjustbox}{max width=\textwidth}
    \begin{tabularx}{\textwidth}{|l|l|X|X|}
        \toprule
        \textbf{User Groups} & \textbf{Intent} & \textbf{Rubric Item} & \textbf{Description} \\
        \midrule
        \multirow{4}{*}{US vs China} & \multirow{4}{*}{Writing Assistance} 
        & Personal Connection and Passion & Western users seek vivid, personal engagement, while Eastern users prefer clear and concise communication, emphasizing empathy and understanding. \\
        \cline{3-4}
        & & Historical and Anecdotal Content & Western users favor detailed historical accounts with personal anecdotes, while Eastern users prefer straightforward summaries with clear information. \\
        \cline{3-4}
        & & Perspective and Tone & Western users prefer second-person perspective, addressing the audience directly, while Eastern users expect the bot to acknowledge and appreciate their contributions. \\
        \cline{3-4}
        & & Refinement in Narrative Style & Western users prefer advanced vocabulary and polished narrative styles, while Eastern users value clarity, conciseness, and brevity. \\
        \hline

        \multirow{4}{*}{US vs China} & \multirow{4}{*}{Creative Content Creation} 
        & Story Continuation & US users prefer detailed and structured script outlines, while Eastern users expect more imaginative and action-packed continuations. \\
        \cline{3-4}
        & & Role-Playing Engagement & US users may expect the assistant to ask for specifications, while Eastern users expect immediate role-play engagement. \\
        \cline{3-4}
        & & Humour and Creative Titles & Both groups enjoy humorous and creative titles, but Eastern users emphasize playful and whimsical text more. \\
        \cline{3-4}
        & & Cultural Resonance and Poetic Elements & Both groups value cultural resonance, but Eastern users place more emphasis on poetic elements. \\
        \hline

        \multirow{3}{*}{US vs India} & \multirow{3}{*}{Writing Assistance} 
        & Acknowledgment and Appreciation & Indian users expect explicit appreciation and acknowledgment of their contributions, while US users do not emphasize this as much. \\
        \cline{3-4}
        & & Personal Connection and Passion & US users prefer vivid engagement with emotions and enthusiasm, while Indian users prioritize shared goals and inclusive language. \\
        \cline{3-4}
        & & Engaging and Descriptive Style & US users prefer engaging and descriptive styles with coherence, while Indian users focus on vivid and friendly tones. \\
        \hline

        \multirow{2}{*}{US vs India} & \multirow{2}{*}{Creative Content Creation} 
        & Story Continuation & US users prefer structured and detailed script outlines, while Indian users expect more imaginative and action-packed stories. \\
        \cline{3-4}
        & & Bedtime Story Personalization & US users expect generic stories, while Indian users prefer more personalized and interactive bedtime storytelling. \\
        \hline
    \end{tabularx}
    \end{adjustbox}

    \caption{Rubric Items Differentiating the Preferences Across User Groups (Separated by Country/Cultural Context) in the domain of Creative Writing and Editing.}
    \label{tab:rubric_culture}
\end{table*}

Many of the results presented in Section~\ref{sec_results} are computed with LLaMA. In Table~\ref{tab:full_gemma_results_bing}, we report \methodCT and \methodFT results for the Gemma model. The results demonstrate that both the \methodCT and \methodFT methods perform well when Gemma is used as the base model.

\begin{table*}[]
\small
    \centering
    \begin{tabular}{lcccc}
        \toprule
        \textbf{Model} & \textbf{LLM Pref (W/L/T)} &  \textbf{LLM conf $\geq$ 75} & \textbf{LLM Pref (W/L/T)} &  \textbf{LLM conf $\geq$ 75} \\
        \midrule
                &  \multicolumn{2}{c}{\textbf{Intent=Programming/Group=Novice}} & \multicolumn{2}{c}{\textbf{Intent=Programming/Group=Expert}}  \\
        \midrule
        \methodCT\ vs Base & 0.6930 / 0.3041 / 0.0029 & 0.5797 / 0.4203 & 0.4489 / 0.5426 / 0.085 & 0.4696 / 0.5304 \\
        \methodCT\ vs Persona & 0.5702 / 0.4298 / 0.0000 & 0.5430 / 0.4570 & 0.5824 / 0.4063 / 0.114 &  0.5924 / 0.4076\\
        \methodCT\ vs Static & 0.5754 / 0.4187 / 0.0058 & 0.5300 / 0.4700 & 0.5625 / 0.4290 / 0.085 &  0.6364 / 0.3636 \\        
        \midrule
        \methodFT\ vs Base & 0.6842 / 0.2953 / 0.0205 & 0.6426 / 0.3574 & 0.5966 / 0.4034/ 0.0000 & 0.5676 / 0.4324\\
        \methodFT\ vs Persona & 0.6439 / 0.4561/ 0.0000 & 0.5290 / 0.4710 & 0.6676 / 0.3295 / 0.0028 & 0.6939 / 0.3061\\
        \methodFT\ vs Static & 0.6433 / 0.3480 / 0.0088 & 0.5807 / 0.4193 & 0.6761 / 0.3210 / 0.0028 & 0.6842 / 0.3158\\
        \midrule
        \methodFT\ vs \methodCT\ & 0.5351 / 0.4649 / 0.0000 & 0.5426 / 0.4574 & 0.6903 / 0.3097 / 0.0000 & 0.6764 / 0.3236 \\
        \bottomrule
    \end{tabular}
    \caption{\methodCT and \methodFT results on the Microsoft Copilot dataset for Gemma for the Novice and Expert groups. 
    The LLM expected confidence $\geq$ 75 is reported and W/L/T=win/lose/tie.}
    \label{tab:full_gemma_results_bing}
\end{table*}

\begin{table*}[]
\small
    \centering
    \begin{tabular}{lcc}
        \toprule
        \textbf{Model} & \textbf{Novice} & \textbf{Expert} \\
        \midrule
        \methodCT\ vs Base & 84.3 / 15.6 & 65.77 / 34.22 \\
        \methodCT\ vs Persona & 73.4 / 26.5 & 61.74 / 38.25 \\
        \methodCT\ vs Static & 76.5 / 23.4 & 55.03 / 44.91 \\     
        \midrule
        \methodFT\ vs Base & 64.51 / 35.48 & 54.12 / 45.88 \\
        \methodFT\ vs Persona & 51.62 / 48.38 & 63.54 / 36.45 \\
        \methodFT\ vs Static & 45.12 / 54.83 & 58.86 / 41.13 \\
        \midrule
        \methodFT\ vs \methodCT\ & 44.22 / 55.78 & 56.16 / 43.84 \\
        \bottomrule
    \end{tabular}
    \caption{Results for Wildchat with LLama. 
    LLM expected confidence (EC) [LLM conf $\geq$ 75] W/L/T=win/lose/tie}
    \label{tab:wildchat_llama}
\end{table*}

Both \methodFT and \methodCT can generate more satisfactory responses than the original DSAT responses.
We compare each of the model responses with the actual DSAT responses from Experts and Novice Samples and prompt GPT4-o to decide which of the response would increase satisfaction level given the conversation history and the followup dissatisfactory user feedback.
Next, we observe whether our methods are chosen over the true dissatisfactory responses and we show results in Table~\ref{tab:normalized_winrates}. We found that compared to baselines, \methodFT and \methodCT provides higher winrates.

\begin{table*}[h]
    \centering
    \begin{tabular}{lccc}
        \toprule
        & Win Rates & Lose Rates & Tie Rates \\
        \midrule
        \methodFT & 0.4848 & 0.2504 & 0.2648 \\
        \methodCT & 0.4877 & 0.2470 & 0.2653 \\
        Static & 0.4283 & 0.3014 & 0.2703 \\
        Persona & 0.4252 & 0.2491 & 0.3257 \\
        Base & 0.4333 & 0.5167 & 0.0500 \\
        \bottomrule
    \end{tabular}
    \caption{Normalized Win, Lose, and Tie Rates}
    \label{tab:normalized_winrates}
\end{table*}

\begin{figure}
    \centering
    \includegraphics[width=0.99\linewidth]{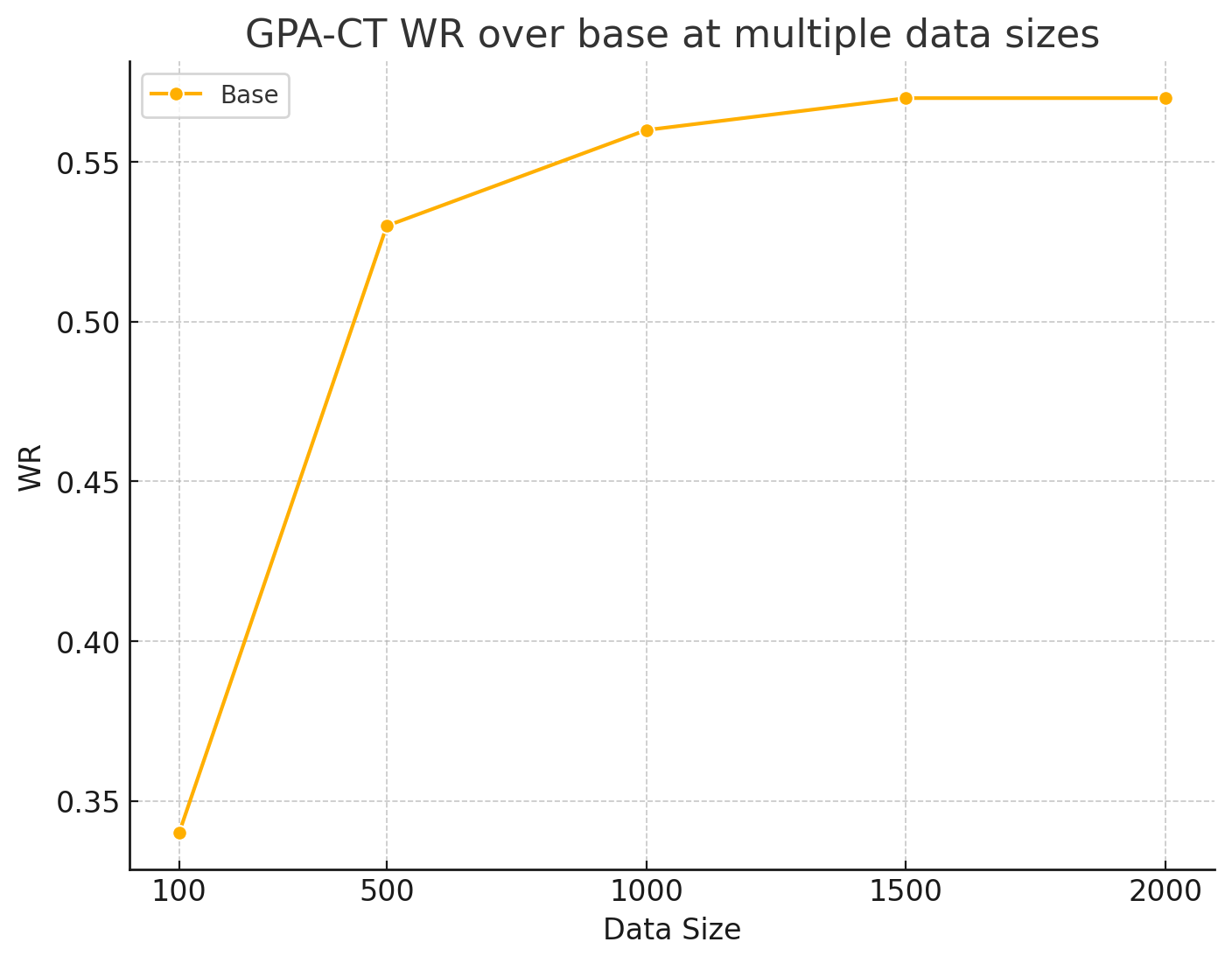}
    \caption{Learning Curve to Show that \methodCT is a data-efficient algorithm. We vary the training data size (100, 500, 1000, 1500, 2000 samples) for extracting our preference rubrics across the Expert and Novice Groups on Wildhat Programming Domain and observe the Win-Rate over the base model on a held-out test set using Prompt~\ref{table:PersonaEvaluation_withEC}. We observe the performance becomes stable using only 1000 examples for rubric creation, making it a data-efficient customization approach.}
    \label{fig:lc}
\end{figure}

In Figure~\ref{fig:lc}, we plot how the training set size used to compute the rubric for \methodCT affects the WinRate (WR). 

\paragraph{DPO with preference data is the optimal choice for \methodFT.}
Table~\ref{tab:choice_of_dpo_vs_otherFT} clearly shows that DPO consistently achieves the highest win rates, making it the best choice for the \methodFT approach. DPO vs. Base achieves a 73.36\% Win Rate, significantly outperforming other methods, demonstrating that DPO leads to stronger preference alignment compared to the base model. Additionally, DPO vs. KTO-Only (70.61\% WR) and DPO vs. KTO-Augmented (61.46\% WR) indicate that DPO still maintains a strong advantage over KTO-based approaches, reinforcing its robustness in fine-tuning.
Rubric augmentation using KTO does help, as KTO-Augmented vs. Base achieves a 67.63\% Win Rate, showing an improvement over KTO vs. Base (53.41\%). However, KTO-based models still fall short of DPO's performance, especially when comparing DPO vs. KTO-Augmented, where DPO wins 61.46\% of the time. This suggests that while rubric augmentation improves alignment, it is not as effective as preference optimization through DPO. Thus, DPO is the optimal choice for \methodFT.

\begin{table}[h]
\small
    \centering
    \begin{tabular}{lcc}
        \toprule
        \textbf{Comparison} & \textbf{Win \%} & \textbf{Lose \%} \\
        \midrule
        DPO vs Base & 73.36\% & 26.64\% \\
        DPO vs KTO-Augmented & 61.46\% & 38.54\% \\
        DPO vs KTO-Only & 70.61\% & 29.39\% \\
        KTO vs Base & 53.41\% & 46.59\% \\
        KTO-Augmented vs Base & 67.63\% & 32.37\% \\
        \bottomrule
    \end{tabular}
    \caption{Win/Loss Percentages of Different Finetuning Methods on Copilot Test Set justifying our best choice of DPO for our remaining $\methodFT$ experiments.}
    \label{tab:choice_of_dpo_vs_otherFT}
\end{table}

\paragraph{\methodshort is generalizable across domain and user-groups.}
Results in Table~\ref{tab:llama_india_us_CT} and Table~\ref{tab:education_domain_performance} provide insights into \methodCT's generalization across different cultural and expertise-based user groups, evaluated at varying EC (confidence) thresholds (65, 70, 75). The results compare GPA-CT against Base, Persona, and Static setups in two domains: India vs. US (Cultural Evaluation) and Education (Novice vs. Expert Evaluation).

\begin{table*}[h]
    \centering
    \begin{tabular}{lcccc}
        \toprule
        \textbf{Model} & \textbf{EC = 65} & \textbf{EC = 70} & \textbf{EC = 75} \\
        \midrule
        \multicolumn{4}{c}{\textbf{Llama-India and US}} \\
        \midrule
        \methodCT\ vs  Base & 0.6057 / 0.3942 & 0.6038 / 0.3961 & 0.6352 / 0.3647  \\
        \methodCT\ vs  Persona & 0.6490 / 0.3509 & 0.6473 / 0.3526 & 0.6666 / 0.3333  \\
        \methodCT\ vs Static & 0.6473 / 0.3526  & 0.7222 / 0.2661 & 0.6666 / 0.3333 \\
        \bottomrule
    \end{tabular}
     \caption{Performance comparison of \methodCT\ models prompt-tuned with India and US Persona at different EC levels}
     \label{tab:llama_india_us_CT}
\end{table*}

\begin{table*}[h]
\centering
\begin{tabular}{lccc}
    \toprule
    \textbf{Model} & \textbf{EC = 65} & \textbf{EC = 70} & \textbf{EC = 75} \\
    \midrule
    \multicolumn{4}{c}{\textbf{Education Domain}} \\
    \midrule
      \methodCT\ vs  Base    & 0.5259 / 0.4740 & 0.5259 / 0.4740  & 0.5229 / 0.4771  \\
    \methodCT\ vs  Persona   & 0.5140 / 0.5259  & 0.5140 / 0.5259  & 0.5271 / 0.5229  \\
    \methodCT\ vs  Static    & 0.5519 / 0.4481 & 0.5519 / 0.4481  & 0.5490 / 0.4599 \\
    \bottomrule
\end{tabular}
\caption{Performance comparison of models in the Education Domain at different EC thresholds.}
\label{tab:education_domain_performance}
\end{table*}

1. Cultural Generalization (India vs. US)
Consistent Gains Over Base: GPA-CT consistently outperforms the Base model across EC thresholds, with a win rate increasing from 60.57\% (EC=65) to 63.52\% (EC=75), showing stable adaptation across cultural contexts.
Stronger Performance Against Persona \& Static: The win rate vs. Persona remains stable ($\approx$65\%), while GPA-CT shows stronger gains vs. Static at EC=70 (72.22\%), indicating that contextual fine-tuning provides advantages over rigid static responses.
Higher EC Improves Differentiation: The lose rate decreases slightly at EC=75, suggesting that higher confidence filtering leads to better generalization.
2. Expertise Generalization (Novice vs. Expert in Education Domain)
Performance Is More Balanced: Compared to cultural evaluation, win rates in the education domain are more balanced across expertise groups, suggesting that novices and experts respond similarly to GPA-CT's responses.
Slight Gains Over Persona: GPA-CT vs. Persona win rate remains around 51-52\%, indicating Persona-based fine-tuning is already well-aligned with education-based responses.
Better Adaptation Over Static: GPA-CT outperforms Static consistently (55.19\% win rate at EC=65 and 54.90\% at EC=75), reinforcing the advantage of dynamic context-aware models over fixed prompts.

GPA-CT generalizes well across cultural differences (India vs. US), maintaining a consistent win rate across EC thresholds.
Higher EC filtering enhances performance differentiation, particularly in cultural evaluations.
Performance in the education domain is more balanced across expertise levels, with smaller gains over Persona but consistent advantages over Static.
EC thresholds affect generalization differently—higher EC benefits cultural evaluations more than expertise-based evaluations.

\subsection{Prompts}
\label{sec_prompts}
The \methodCT and \methodFT methods use a number of LLM prompts which we describe in this appendix.
User preferences are inferred for both satisfaction and dissatisfaction judgments. The prompt for inferring user preference for satisfaction in line 8 of Algorithm~\ref{algo:rubricextraction} can be found in the prompt in Table~\ref{table:sat_InferUserExpectation} titled \texttt{LLM.InferUserPreference} (for SAT judgment).
Similarly, inferring the user preference in line 10 of Algorithm~\ref{algo:rubricextraction} under the dissatisfaction condition can be accomplished using the prompt titled \texttt{LLM.InferUserPreference} (for DSAT judgment).
The aspects and likert rating are computed using the  \texttt{LLM.ExtractAspectsAndLikert} prompt in Table~\ref{table:computelikert}.
The \texttt{LLM.ModifyPromptWithRubrics} in Table~\ref{table:ModifyPrompt} is used to generate a response from the LLM. This prompt is used for \methodCT inference in Algorithm~\ref{algo:context-tuning} and \texttt{Rubric-Guided Data Generation} in Algorithm~\ref{algo:data-generation}.
The \textit{Persona} results are generated with the \texttt{LLM.PersonaEvaluation} prompt presented in  Table~\ref{table:PersonaEvaluation_withoutEC} and Table~\ref{table:PersonaEvaluation_withEC}.
The win-lose-tie rates for the results presented in Section~\ref{sec_results} are computed from the prompt in Table~\ref{table:PersonaEvaluation_withoutEC}. 
Table~\ref{table:ind_dsat_evaluation} provides the \texttt{LLM.IndividualDSATEvaluation}  prompt.
Finally, labeling a user's expertise in a conversation is accomplished using the \texttt{LLM.ExpertiseLabellingPrompt} prompt in Table~\ref{table:expertise}.

\begin{table*}[h!]
\centering
\begin{tabular}{>{\raggedright\arraybackslash}p{0.95\linewidth}}
\hline
\textbf{LLM.InferUserPreference (for SAT judgment)} \\
\hline
\# OVERVIEW \\

You will be given a conversation between a User and an AI agent. Your task is to assess the reasons of user's happiness based on the conversation history and the bot response.
\\
\# TASK: 
\\

Classify the user's intent from the conversation \{conversation history\}. Also determine what the user expects from the bot and why the user finds the bot's response \{user remarks\} useful. Determine based on whatever the user remarks after the bot's response \{user remarks\}.
\\

\# ANSWER FORMAT 

Format your output as JSON Object where the keys are user-intent, user-expectation-from-bot and reasons-for-happiness. Do not output anything else except this. \\
\hline
\end{tabular}
\caption{LLM.InferUserPreference (for SAT judgment)}
\label{table:sat_InferUserExpectation}
\end{table*}

\begin{table*}[h!]
\centering
\begin{tabular}{>{\raggedright\arraybackslash}p{0.95\linewidth}}
\hline
\textbf{LLM.InferUserPreference (for DSAT judgment)} \\
\hline
 \# OVERVIEW \\

You will be given a conversation between a User and an AI agent. Your task is to assess the reasons of user's frustration based on the conversation history and the bot response.
\\
\# TASK: 
\\

Classify the user's intent from the conversation \{conversation history\}. Also determine what the user expects from the bot and why the user finds the bot's response \{user remarks\} frustrating. Determine based on whatever the user remarks after the bot's response \{user remarks\}.
\\

\# ANSWER FORMAT 

Format your output as JSON Object where the keys are user-intent, user-expectation-from-bot and reasons-for-frustration. Do not output anything else except this. \\
\hline
\end{tabular}
\caption{LLM.InferUserPreference (for DSAT judgment)}
\label{table:dsat_InferUserExpectation}
\end{table*}

\begin{table*}[h!]
\centering
\begin{tabular}{>{\raggedright\arraybackslash}p{0.95\linewidth}}
\hline
\textbf{LLM.ExtractAspectsAndLikert} \\
\hline
\# OVERVIEW \\

\# Task Overview: 

You have to compare preference explanations of two user groups based on some aspects in \{intent-name\}, and provide ratings of 1-5 depending on how much different are their preference from the bot while interaction. 
You have to update the comparison output based on what was observed previously \{previous-aspects\} and the current observed differences in preference explanations between group 1 and group 2 described below. Make sure that if there is no observed datapoints for an aspect in either group, provide the least rating in that case. \\

\# Primary Intent \\

Intent : \{intent\}

\# Preference explanations of Group 1

\{preferences-of-group 1\}

\# Preference explanations of Group 2

\{preference-of-group 2\} \\

\# Annotation Guidelines on a scale of 1-5

 1 : It indicates there is no observed difference between the preferences of two groups on this aspect,
 
 2 : It indicates there is a minor difference between the preferences of two groups on this aspect,
 
 3 : It indicates moderate difference between the preferences of two groups on this aspect,
 
 4 : It indicates remarkable difference between the preferences of two groups on this aspect, 
 
 5 : It indicates undoubetedly stark difference between the preferences of two groups on this aspect.
 
\# Output Format 

Format your output as JSON where keys are aspects and values are 1) ratings from 1-5 and 2) Interpretation of the rating in 2-3 sentences. \\
\hline
\end{tabular}
\caption{LLM.ExtractAspectsAndLikert}
\label{table:computelikert}
\end{table*}

\begin{table*}[h!]
\centering
\begin{tabular}{>{\raggedright\arraybackslash}p{0.95\linewidth}}
\hline
\textbf{LLM.ModifyPromptWithRubrics} \\
\hline
\# OVERVIEW \\

\# Task

You will be provided with a conversation between a user and bot. Based on the conversation history, you have to generate a suitable response. Make sure that you follow some rules while generating the response.

\# Conversation History

\{conversation-history\}

\# User Input 

\{user-input\}

\# Some Rules to Follow

\{rubrics-for-intent-and-group\}

\# Output Format

Format your output as a JSON Object with response as key. Do not output anything else except this JSON.
\\
\hline

\end{tabular}
\caption{LLM.ModifyPromptWithRubrics}
\label{table:ModifyPrompt}
\end{table*}

\begin{table*}[h!]
\centering
\begin{tabular}{>{\raggedright\arraybackslash}p{0.95\linewidth}}
\hline
\textbf{LLM.ExpertiseLabeling} \\
\hline
\# OVERVIEW \\

\# OVERVIEW 

You will be given a conversation history between a User and an AI agent. Your task is to determine user's expertise in the subject of the conversation.

\# USER EXPERTISE 

User expertise levels in a conversation subject range from novice, indicating a lack of familiarity with fundamental concepts, to expert or master, denoting a deep understanding of relevant vocabulary, concepts, and principles. 

- Novice: A subject novice is a person who has little or no familiarity with a specific topic or domain. A subject novice may ask questions that are vague, general, irrelevant, or based on incorrect assumptions. A subject novice may also have difficulty understanding the terminology, concepts, or arguments of experts or more knowledgeable people in the subject. They may ask basic or general questions that can be answered by simple definitions, examples, or facts. They may not be aware of the sources, methods, concepts, or terminology that are relevant to the subject.  

- Intermediate: A subject intermediate is someone who has some basic knowledge or familiarity with a certain topic, but not enough to be considered an expert or a novice. A subject intermediate can ask general questions that reflect their curiosity or interest in the topic, but not very specific or complex ones that require deeper understanding or analysis. A subject intermediate might have learned some terms or concepts related to the topic, but not how to apply them in different contexts or situations.  

- Expert: A subject expert is someone who can apply relevant concepts and terminology to different scenarios and problems. They can analyze and interpret data, compare and contrast different methods or approaches, and justify their reasoning with evidence. The user also demonstrates curiosity and interest in the subject by asking questions that go beyond the surface level and explore the deeper implications and connections of the topic. He has a deep and comprehensive understanding of a specific topic or field, and can use specialized terms and references to communicate their knowledge. A subject expert can state accurate facts, provide relevant examples, and cite authoritative sources related to their topic or field.  
 
- Unknown: There is not enough information to determine the user's expertise.

\#\# OUTPUT FORMAT \#\# 

Format your output as JSON Object with key as Expertise-label and values as either Novice, Intermediate, Expert or Unknown.

\#\# INPUT \#\# 

Conversation History

\#\# OUTPUT \#\# 
\\
\hline
\end{tabular}
\caption{Prompt to Classify Expertise}
\label{table:expertise}
\end{table*}

\begin{table*}[h!]
\centering
\begin{tabular}{>{\raggedright\arraybackslash}p{0.95\linewidth}}
\hline
\textbf{LLM.PersonaRole-Playing} \\
\hline
\# OVERVIEW \\

You will be given a conversation between a User and an AI agent. Your task is to generate response that would tailor to a  \{group\} user in the \{intent.domain\} domain.
\\
\hline
\end{tabular}
\caption{Persona-Role Playing Response Generation}
\label{table:persona-aware}
\end{table*}

\begin{table*}[h!]
\centering
\begin{tabular}{>{\raggedright\arraybackslash}p{0.95\linewidth}}
\hline
\textbf{LLM.StaticPrompting} \\
\hline
\# Overview \\

Tell me the expectations of a \{group\} user in a \{intent.domain\} domain from the chatbot. Answer in a few sentences.
\\
\hline
\end{tabular}
\caption{Static}
\label{table:static}
\end{table*}

\begin{table*}[h!]
\centering
\begin{tabular}{>{\raggedright\arraybackslash}p{0.95\linewidth}}
\hline
\textbf{LLM.PersonaEvaluation} \\
\hline
\# OVERVIEW

\# Task

Imagine yourself as a {group} user in the \{intent.domain\} domain. Based on your persona and the conversation history, you have to judge which response would you prefer among Option A and Option B along with the step-by-step reasoning. \\

\# Conversation History \\
\{conversation-history\}

\# Option A \\
{option1} \\

\# Option B \\
{option2} \\

\# Output Format \\
Format your output as a JSON Object with key as Reason and Output. Output the step-by-step reasoning and then Option A or Option B or can't decide. You should not output anything except the JSON.
\\
\hline
\end{tabular}
\caption{Prompt used for LLM-as-a-Personalized-Judge}
\label{table:PersonaEvaluation_withoutEC}
\end{table*}

\begin{table*}[h!]
\centering
\begin{tabular}{>{\raggedright\arraybackslash}p{0.95\linewidth}}
\hline
\textbf{LLM.PersonaEvaluationwithEC} \\
\hline
\# OVERVIEW

\# Task

Imagine yourself as a \{group\} user in the \{intent.domain\} domain. Based on your persona and the conversation history, you have to judge which response would you prefer among Option A and Option B along with the step-by-step reasoning. \\

Additionally, assess your confidence in this decision by assigning a certainty level from 1 to 100. Use the following guidelines to assign the certainty level:   \\
1–20 (Uncertain): The user profile provides insufficient or minimal evidence. The decision is
largely based on weak or indirect hints.  \\
21–40 (Moderately Confident): There is noticeable evidence supporting a preference, though it is
not comprehensive, and other interpretations are possible.
41–60 (Quite Confident): You find clear and convincing evidence that supports your prediction,
though it is not entirely decisive.  \\
61–80 (Confident): The user profile contains strong evidence that clearly supports your prediction,
with very little ambiguity. \\
81–100 (Highly Confident): The user profile provides direct and explicit evidence that decisively
supports your prediction

\# Conversation History \\
\{conversation-history\}

\# Option A \\
{option1} \\

\# Option B \\
{option2} \\

\# Output Format \\
Format your output as a JSON Object with keys as Reason, Output, Confidence. Output the step-by-step reasoning and then Option A or Option B and the confidence value from 1-100. You should not output anything except the JSON. \\
\hline
\end{tabular}
\caption{Prompt used for LLM-as-a-Personalized-Judge with confidence \cite{dong-etal-2024-llm}}
\label{table:PersonaEvaluation_withEC}
\end{table*}

\begin{table*}[h!]
\centering
\begin{tabular}{>{\raggedright\arraybackslash}p{0.95\linewidth}}
\hline
\textbf{LLM.IndividualDSATEvaluation} \\
\hline
\# OVERVIEW \\
\# Task

In the conversation context between user and assistant: \{conversation\_history\}, based on user utterance : \{user\_utterance\}, when the bot responds : \{{bot}\_{response}\}, the user felt \{judgment\_label\}, then he provides a feedback by commenting \{{feedback}\_{comment}\}. 

You have to compare Option A and Option B and judge which response is very different from reference bot response \{{bot}\_{response}\}, such that the user will not provide a followup comment \{{feedback}\_{comment}\}.

\# Option A

\{optionA\}

\# Option B

\{optionB\}

\# Output Format

Format your output as a JSON Object with keys as Option and reasoning. Output either Option A or Option B or can't decide. You should not output anything except the JSON. Do not judge based on user expertise. Judge only based on which response is very different from reference bot response \{bot\_{response}\}.
\\
\hline
\end{tabular}
\caption{Individual DSAT Evaluation}
\label{table:ind_dsat_evaluation}
\end{table*}

\end{document}